\documentclass{article} 
\PassOptionsToPackage{numbers,compress}{natbib}
\usepackage[preprint]{style/arxiv}

\usepackage{geometry}
\usepackage[most]{tcolorbox}
\usepackage{amsmath,amssymb,amsthm,amsfonts,bm}
\usepackage{booktabs}
\usepackage{float}
\usepackage{array}
\usepackage{colortbl}
\usepackage{mathtools}
\usepackage{thmtools}
\usepackage{booktabs}
\usepackage{xfrac}
\usepackage{xspace}
\usepackage{graphicx}
\usepackage[normalem]{ulem}
\usepackage{subfigure}
\usepackage{todonotes}
\usepackage[toc,page,header]{appendix}
\usepackage{minitoc}
\usepackage{url}
\usepackage{comment}
\usepackage{algorithm} 
\usepackage{wrapfig}
\usepackage{multirow}
\usepackage{multicol}
\usepackage[T1]{fontenc}
\usepackage{hyperref}
\usepackage{url}
\usepackage{cleveref}
\usepackage{enumitem}
\usepackage{soul}
\setuldepth{Test}
\usepackage{placeins}
\usepackage[noend]{algpseudocode}
\usepackage{siunitx}
\creflabelformat{equation}{#2\textup{#1}#3}

\usepackage{caption}
\usepackage{amsmath,amsfonts,bm}

\def\eqref#1{equation~\ref{#1}}

\def\1{\bm{1}}

\DeclareMathAlphabet{\mathsfit}{\encodingdefault}{\sfdefault}{m}{sl}
\SetMathAlphabet{\mathsfit}{bold}{\encodingdefault}{\sfdefault}{bx}{n}

\usepackage{todonotes}
\usepackage{graphicx}  
\usepackage{adjustbox}
\usepackage{ifthen}

\usepackage[normalem]{ulem}
\usepackage{xcolor}

\newcommand{\scrcmd}{\textsuperscript}
\newcommand{\scr}[1]{\scrcmd{#1}}

\title{Void in Language Models}

\author{
  Mani Shemiranifar \\
  Independent Researcher \\
  m.shemiranifar@gmail.com
}

\begin{document}
\setcounter{tocdepth}{2}
\doparttoc 
\renewcommand\ptctitle{}
\faketableofcontents 

\maketitle

\begin{abstract}
Despite advances in transformer-based language models (LMs), a fundamental question remains largely unanswered: Are all layers activated during inference? We investigate this question by detecting unactivated layers (which we refer to as \textbf{Voids}) using a non-trainable and parameter-free adaptive computation method called L2 Adaptive Computation (LAC). We adapt LAC from its original efficiency-focused application to trace activated layers during inference. This method monitors changes in the L2-norm of activations to identify voids. We analyze layer activation in instruction-tuned LMs across two phases: Prompt Processing (PP), where we trace activated layers for each token in the input prompts, and Response Generation (RG), where we trace activated layers for each generated token. We further demonstrate that distinct layers are activated during these two phases. To show the effectiveness of our method, we evaluated three distinct instruction-tuned LMs from the Llama, Mistral, and Qwen families on three benchmarks: MMLU, GPQA Diamond, and BoolQ. For example, on MMLU with a zero-shot setting, skipping voids in Qwen2.5-7B-Instruct resulted in an improvement from 69.24 to \textbf{71.29} while the model uses only \textbf{30\%} of the layers. Similarly, Mistral-7B-Instruct-v0.3 on GPQA Diamond improved from 13.88 to \textbf{18.36} when using \textbf{70\%} of the layers during both the PP and RG phases. These results show that not all layers contribute equally during inference, and that selectively skipping most of them can improve the performance of models on certain tasks.
\footnote{The code is available in \url{https://github.com/manishemirani/void_in_language_models}}
\end{abstract}

\section{Introduction}
In recent years, there have been remarkable advances in transformer-based language models~\citep{vaswani2017attention}, broadening their capabilities from natural language processing tasks to complex reasoning problems~\citep{openai2024openaio1card, deepseekai2025deepseekr1incentivizingreasoningcapability}. However, despite these achievements, these models remain largely black boxes. Several approaches have been developed to make LMs more transparent, including circuit analysis~\citep{elhage2021mathematical, geva2020transformer}, activation analysis~\citep{gurnee2023finding, burns2022discovering}, probing~\citep{alain2018understandingintermediatelayersusing, orgad2024llms}, and mechanistic interpretability~\citep{wang2022interpretability, elhage2022solu, bricken2023monosemanticity}.

In addition to the methods mentioned above, analyzing layer activation patterns in Language Models (LMs) offers valuable insights into how these models process inputs of varying complexity. We can pinpoint which layers get activated during the inference by employing Adaptive Computation techniques. These methods were originally designed to improve the efficiency~\citep{dehghani2022efficiencymisnomer} of neural networks~\citep{bengio2016conditionalcomputationneuralnetworks}, recurrent neural networks~\citep{graves2017adaptivecomputationtimerecurrent, banino2021pondernetlearningponder}, language models~\citep{dehghani2019universaltransformers, schuster2022confidentadaptivelanguagemodeling, jiang2024dllm}, and vision models~\citep{bolukbasi2017adaptiveneuralnetworksefficient, wang2018skipnetlearningdynamicrouting, xue2023adaptivecomputationelasticinput}. However, these techniques are not straightforward to use when analyzing layer activation patterns in LMs, since they need to be trained specifically for this purpose and often lead to degraded model performance.

To address these limitations, we utilize L2 Adaptive Computation (LAC)~\citep{shemiranifar2023l}, which offers a simple approach compared to existing methods. LAC operates without requiring any training, making it particularly well-suited for analyzing LMs. This method functions by monitoring changes in the L2-norm of layer activations to identify halting points. LAC provides three distinct halting stages: per-batch, per-example, and per-token. We focus on the per-token setting since we aim to observe which layers are activated while LMs process and generate tokens

\newpage

Our analysis examines layer activation patterns across two distinct phases:
\begin{itemize}
    \item \textbf{Prompt Processing (PP)}: We use LAC's per-token setting to track activated layers as the model processes each token within the input prompts.
    \item \textbf{Response Generation (RG)}: We use LAC's per-token setting to track activated layers as the model generates each token based on the given input prompt.
\end{itemize}

We evaluate our method on three distinct instruction-tuned Language Models from the Llama, Mistral, and Qwen families, across a range of challenging benchmarks: MMLU, GPQA Diamond, and BoolQ. Our findings reveal that Language Models do not consistently activate all layers during inference. Surprisingly, we demonstrate that selectively skipping these detected Voids can lead to performance improvements on certain tasks, even when utilizing a significantly reduced subset of the model's total layers. 

For instance, on MMLU with a zero-shot setting, skipping Voids in Qwen2.5-7B-Instruct resulted in an improvement from 69.24 to \textbf{71.29} accuracy while using only \textbf{30\%} of the model's layers. Similarly, Mistral-7B-Instruct-v0.3 on GPQA Diamond saw an improvement from 13.88 to \textbf{18.36} accuracy when using approximately \textbf{70\%} of its layers across both PP and RG phases. Furthermore, we demonstrate that distinct layers are activated during the Prompt Processing and Response Generation phases, highlighting task-dependent computational allocation within these models.

The consistent presence of Voids across different models and tasks, coupled with the observed performance gains when selectively skipping them, strongly indicates that layers contribute unequally during inference. Our method, which precisely identifies these Voids, thus provides valuable insights into the inner workings of Language Models. This approach opens new avenues for interpretability, targeted knowledge editing, and hallucination detection by pinpointing layers that are less crucial for specific computations.
\section{Background}
In this section, we will review the core concept of L2 adaptive computation, detailing how measurements of the L2-norm (Section \ref{ref:L2}) can be applied at various levels of granularity. Next, we examine LAC's halting mechanism (Section \ref{ref:Halt}), which leverages these measurements to halt computations. We then analyze the threshold sensitivity (Section \ref{ref:Threshold}). Finally, we discuss the practical implementation of granular halting through masking (Section \ref{ref:masking}).

\subsection{L2 Adaptive Computation}
\label{ref:L2}
L2 Adaptive Computation (LAC) considers the model's progress as a halting criterion, where this algorithm defines the progress based on the changes in the L2-norm made by each layer applied sequentially. LAC requires a step function to perform; consider $S_{t}(h_{t}) = h_{t + 1}$ for $\mathbf{t} = \{1, ..., T\}$ where $T$ represents the maximum number of layers and $h \in \mathbb{R}^{B \times L \times D}$ denoting the hidden state at step $t$ with $B$, $L$ and $D$ representing batch size, length, and depth, respectively. LAC offers three distinct levels for halting the computational steps:

\begin{enumerate}
    \item Per-Batch: In this level, the L2-norm will be calculated for the whole given batch and accordingly halt the computation for this level:
    
    \begin{equation}
        \|{h}\|_2 = \sqrt{\Sigma_{i=1}^{b}\Sigma_{j=1}^{l}\Sigma_{k=1}^{d} h_{ijk}^2} \quad \|{h}\|_2 \in \mathbb{R}^{1}
    \end{equation}

    \item Per-Example: In this level, L2-norm will be calculated for each example in a batch, and the halting mechanism will stop computation for each example separately:
    
    \begin{equation}
        \|{h}\|_2 = \{\sqrt{\Sigma_{j=1}^{l}\Sigma_{k=1}^{d} h_{1jk}^2}, \sqrt{\Sigma_{j=1}^{l}\Sigma_{k=1}^{d} h_{2jk}^2}, ..., \sqrt{\Sigma_{j=1}^{l}\Sigma_{k=1}^{d} h_{bjk}^2}\} \quad \|{h}\|_2 \in \mathbb{R}^{B \times 1}
    \end{equation}
    \item Per-Token: In this level. L2-norm will be calculated for each token in examples, therefore, the halting mechanism stops the computation for each token individually:
    
    \begin{equation}
        \|{h}\|_2 = \{\sqrt{\Sigma_{k=1}^{d} h_{11k}^2}, \sqrt{\Sigma_{k=1}^{d} h_{12k}^2}, ..., \sqrt{\Sigma_{k=1}^{d} h_{blk}^2}\} \quad \|{h}\|_2 \in \mathbb{R}^{B \times L \times 1}
    \end{equation}
\end{enumerate}

\subsection{LAC's Halting Mechanism}
\label{ref:Halt}
Having defined how the L2-norms are calculated at various granularities (Section \ref{ref:L2}), we now describe how these measurements are used to decide whether to halt computation for a given input (batch, example, or token). The halting decision is based on the model's "progress" during each computational step $t$. Progress at step $t$ is defined as the change in the L2-norm of the hidden state from step $t-1$ to $t$:

\begin{equation}
    \label{eq:delta}
    \delta_{t} = \|{h_{t}}\|_2 - \|{h_{t-1}}\|_2
\end{equation}

where $\delta_{t}$ denotes the progress of the model at step $t$. Depending on how to compute the L2-norm (\ref{ref:L2}), the shape of $\delta$
changes, e.g., $\delta \in \mathbb{R}^{1}$ for the per-batch setting, $\delta \in \mathbb{R}^{b \times 1}$ for the per-example setting and $\delta \in \mathbb{R}^{b \times l \times 1}$ for the per token setting. LAC keeps track of all progress measurements from the first step to the last step($t$) before halting computations:

\begin{equation}
    \Delta_t = \{\delta_{1}, \delta_{2} \ldots, \delta_{t}\}
\end{equation}

Then $\Delta_t$ is used to calculate a dynamic threshold $\lambda_t$ at step $t$:

\begin{equation}
    \label{eq:lambda}
    \lambda_{t} = \alpha \lvert \max(\Delta_t) - \min(\Delta_t) \rvert \quad
\end{equation}

Here, $\lambda_t$ represents the minimum required change in activation (progress) for the activations to continue to step $t+1$. The parameter $\alpha \in (0, 1]$ acts as a knob to control the decisiveness of the threshold.

LAC halts the computation for a specific unit (batch, example, or token) before executing step $t+1$ if the progress made at step $t$ falls below the calculated threshold $\lambda_t$ for that unit: $\delta_{t} < \lambda_t$. This condition is evaluated independently for each unit at the chosen granularity.

\subsection{Threshold Sensitivity}
\label{ref:Threshold}
The $\alpha$ parameter in Equation \ref{eq:lambda} directly controls the threshold sensitivity. A lower value of $\alpha$ results in a smaller $\lambda_t$, creating a more permissive threshold that allows more computational steps. Conversely, a higher $\alpha$ results in a larger $\lambda_t$, leading to a more decisive threshold and fewer steps. The LAC framework distinguishes between $\alpha_t$ used during training and $\alpha_e$ used during evaluation. The optimal $\alpha$ value depends on factors like the complexity of the input data. For our analysis, we use $\alpha_e$ to control the halting behavior of pre-trained models during inference.

\subsection{Masking}
\label{ref:masking}

LAC can dynamically halt computations at different granularities: for the entire batch, for individual examples within the batch, or for individual tokens within examples. Per-batch halting is relatively straightforward since the computation for the entire batch must proceed until the halting condition is met for all active examples; therefore, it is possible to halt the computations. However, implementing per-example and per-token halting efficiently on hardware accelerators is challenging due to poor performance with arbitrary sparse operations (i.e., skipping computations for arbitrary examples or tokens).

Therefore, LAC employs masking to achieve this granular halting mechanism without requiring highly sparse operations. We define $\delta_{E}$ and $\lambda_{E}$ as the change in the L2-norm and the corresponding threshold for examples, respectively. Similarly, $\delta_{T}$ and $\lambda_{T}$ are defined for tokens.

In per-example halting, masking is used to zero out the activations for an entire example when its specific halting condition is met. Let $h_i$ represent the activation corresponding to the $i$-th example in the batch. The masking operation is defined as follows:

\begin{align}
    h_i = \begin{cases} 0 & \text{if } \delta_{E}^i < \lambda_{E} \\ h_i & \text{otherwise} \end{cases} \quad \forall \; 1 \leq i \leq l
\end{align}

where $0$ is a zero tensor with the same shape as $h_i$, and $l$ is the maximum number of examples in the batch. When condition $\delta_{E}^i < \lambda_{E}$ is met, for example, $i$, its activations $h_i$ are set to zero, effectively halting further computation for this example in subsequent layers.

Per-token requires a finer-grained masking approach. Here, activations corresponding to individual tokens are zeroed out if their specific halting condition is met. Let $h_{ij}$ represent the activation for the $j$-th token within the $i$-th example. The masking operation is defined as:

\begin{align}
    h_{ij} = \begin{cases} 0 & \text{if } \delta_{T}^{ij} < \lambda_{T} \\ h_{ij} & \text{otherwise} \end{cases} \quad \forall \; 1 \leq i \leq l, \; 1 \leq j \leq N
\end{align}

Here, $0$ represents a zero tensor of the same dimension as $h_{ij}$, $l$ is the maximum number of examples, and $N$ is the maximum sequence length in the batch. When condition $\delta_{T}^{ij} < \lambda_{T}$ is met for a specific token, its corresponding activations $h_{ij}$ are set to zero, halting further processing for that particular token position within the example.
\section{Detecting and Skipping Voids in LMs}
In this section, we modify the LAC threshold formula to precisely identify activated layers (voids) during LM inference. We then demonstrate the impact of selectively skipping these identified voids on the performance of three distinct instruction-tuned LMs across three benchmarks. Furthermore, we analyze the effect of the parameter $\alpha$ on layer usage and model performance when skipping voids, focusing on the BoolQ\citep{clark2019boolq} dataset. Finally, we delve into the specific layer usage patterns and the underlying L2-norm dynamics that characterize voids.

\subsection{Experimental Setup}
For our experiments, we considered three instruction-tuned LMs, Qwen2.5-7B-Instruct~\citep{qwen2025qwen25technicalreport}, Llama3-8B-Instruct~\citep{grattafiori2024llama3herdmodels}, and
Mistral-7B-Instruct-v0.3~\citep{jiang2023mistral7b}. For our evals, we used Massive Multitask Language Understanding (MMLU)~\citep{hendrycks2021measuringmassivemultitasklanguage}, a common world-knowledge and problem-solving benchmark, Graduate-Level Google-Proof Q\&A (GPQA Diamond) ~\citep{rein2023gpqagraduatelevelgoogleproofqa}, a graduate-level multiple-choice benchmark in the biology, physics, and chemistry domains, and BoolQ~\citep{clark2019boolq}, a common yes/no reading comprehension benchmark. We evaluated these models in two configurations for each benchmark. First, we evaluated the LMs using their full layer capacity (i.e., without skipping any layers). Second, we evaluated the LMs when selectively skipping the detected unactivated layers (voids) using our modified LAC method. For the main performance evaluation across all benchmarks, we set the $\alpha$ parameter to $0.8$. However, for the specific analysis of $\alpha$'s effect on the BoolQ benchmark, we explored a range of $\alpha$ values from 0.1 to 1.0 to illustrate how varying the threshold impacts the amount of skipped layers and model behavior. All experiments were performed using a single 24GB RTX 3090 GPU.

\subsection{Modifying LAC}
To precisely identify unactivated layers (voids) during LM inference, we modified the standard LAC threshold formula (Section \ref{eq:lambda}) by removing the absolute value:

\begin{equation}
    \lambda_{t} = \alpha (\max(\Delta_t) - \min(\Delta_t)) \quad
\end{equation}

The original LAC formulation used the absolute value to compute a threshold aimed at efficiently halting computations during training. In contrast, for the purpose of detecting voids at inference time in pre-trained models, we found that removing the absolute value allows the threshold $\lambda_t$ to better capture the subtle changes in L2-norm that indicate when a layer's contribution falls below a significant level, thereby identifying it as a void.

\subsection{Skipping Voids}
\label{sec:skip}
\begin{figure*}[!t]
    \centering
    \includegraphics[width=1.0\textwidth]{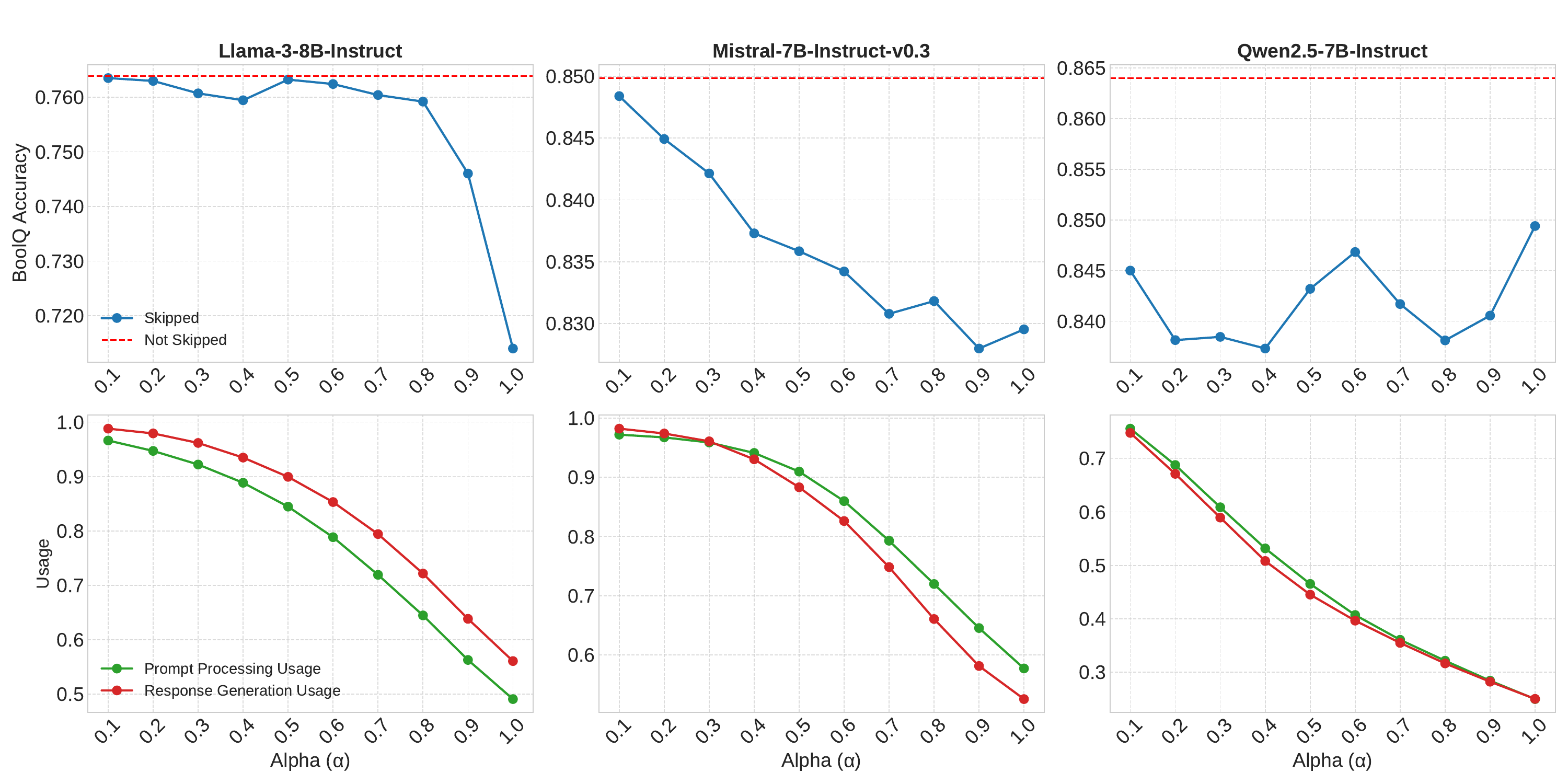}
    \caption{BoolQ accuracy and Percentage of total layers used (relative to total model layers) as a function of $\alpha$. \emph{Left:} Llama-3-8B-Instruct; \emph{Middle:} Mistral-7B-Instruct-v0.3; \emph{Right:} Qwen2.5-7B-Instruct. In the top row, the blue line shows performance when skipping voids, and the red horizontal line indicates the full model's performance. In the bottom row, the red line represents average layer usage during Response Generation (RG), and the green line shows average layer usage during Prompt Processing (PP). Usage is shown as a percentage of total model layers.}
    \label{fig:boolq}
\end{figure*}
Table \ref{tab:performance} presents the performance scores of the LMs on the three benchmarks when using their full layer capacity ("Not Skipped") compared to when selectively skipping the detected voids ("Skipped"), using $\alpha=0.8$. The table also shows the average percentage of layers utilized per token during the Prompt Processing (PP) and Response Generation (RG) phases when skipping is enabled.

\begin{table}[H]
  \centering
  \scriptsize 
  \setlength{\tabcolsep}{3pt} 
  \resizebox{\textwidth}{!}{ 
    \begin{tabular}{l*{6}{S[table-format=2.2]}}
      \toprule
      \multirow{2}{*}{\textbf{Model}} & 
      \multicolumn{2}{c}{\textbf{MMLU$_{zero-shot}$}} & 
      \multicolumn{2}{c}{\textbf{GPQA Diamond}} & 
      \multicolumn{2}{c}{\textbf{BoolQ}} \\
      \cmidrule(lr){2-3} \cmidrule(lr){4-5} \cmidrule(lr){6-7}
      & {\textbf{Not Skipped}} & {\textbf{Skipped}} & {\textbf{Not Skipped}} & {\textbf{Skipped}} & {\textbf{Not Skipped}} & {\textbf{Skipped}} \\
      \midrule
      Llama-3-8B-Instruct & \textbf{61.18} & 60.42 & 29.11 & \textbf{30.53} & \textbf{76.38} & 75.92 \\
      Qwen2.5-7B-Instruct & 69.24 & \textbf{71.29} & \textbf{34.78} & 33.33 & \textbf{86.40} & 83.81 \\
      Mistral-7B-Instruct-v0.3 & \textbf{59.70} & 59.29 & 13.88 & \textbf{18.36} & \textbf{84.98} & 83.18 \\
      \midrule
      \multirow{2}{*}{\textbf{Usage}} & 
      \multicolumn{2}{c}{\textbf{MMLU$_{zero-shot}$}} & 
      \multicolumn{2}{c}{\textbf{GPQA Diamond}} & 
      \multicolumn{2}{c}{\textbf{BoolQ}} \\
      \cmidrule(lr){2-3} \cmidrule(lr){4-5} \cmidrule(lr){6-7}
      & {\textbf{PP}} & {\textbf{RG}} & {\textbf{PP}} & {\textbf{RG}} & {\textbf{PP}} & {\textbf{RG}} \\
      \midrule
      Llama-3-8B-Instruct(Layers=32) & 0.53 & 0.43 & 0.62 & 0.65 & 0.64 & 0.72 \\
      Qwen2.5-7B-Instruct(Layers=28) & 0.29 & 0.30 & 0.32 & 0.31 & 0.36 & 0.36 \\
      Mistral-7B-Instruct-v0.3(Layers=32) & 0.71 & 0.72 & 0.74 & 0.74 & 0.71 & 0.66 \\
      \bottomrule
    \end{tabular}
  }
  \caption{Evaluation scores and average percentage of layers used when skipping voids for the three LMs on MMLU, GPQA Diamond, and BoolQ benchmarks in a zero-shot setting, with $\alpha=0.8$. Results are shown for full model capacity ("Not Skipped") and when selectively skipping voids ("Skipped"). Average layer usage is presented separately for the Prompt Processing (PP) and Response Generation (RG) phases, normalized by the total number of layers and shown as a percentage.}
  \label{tab:performance}
\end{table}

Notably, we observe performance gains on some benchmarks despite skipping a significant proportion of layers. For example, Qwen2.5-7B-Instruct shows a 2.05 percentage point improvement in the MMLU zero-shot score (from 69.24 to 71.29) while using only approximately 30\% of its total layer capacity (average of 29\% for PP and 30\% for RG). Similarly, Mistral-7B-Instruct-v0.3 achieves a performance gain of 4.48 percentage points on GPQA Diamond (from 13.88 to 18.36) while utilizing approximately 74\% of its layers during both phases. These results demonstrate that selective skipping of voids, identified by LAC, can lead to performance improvements on certain tasks, even when significantly reducing the computation per token.

\begin{figure*}[!t]
    \centering
    \includegraphics[width=1.0\textwidth]{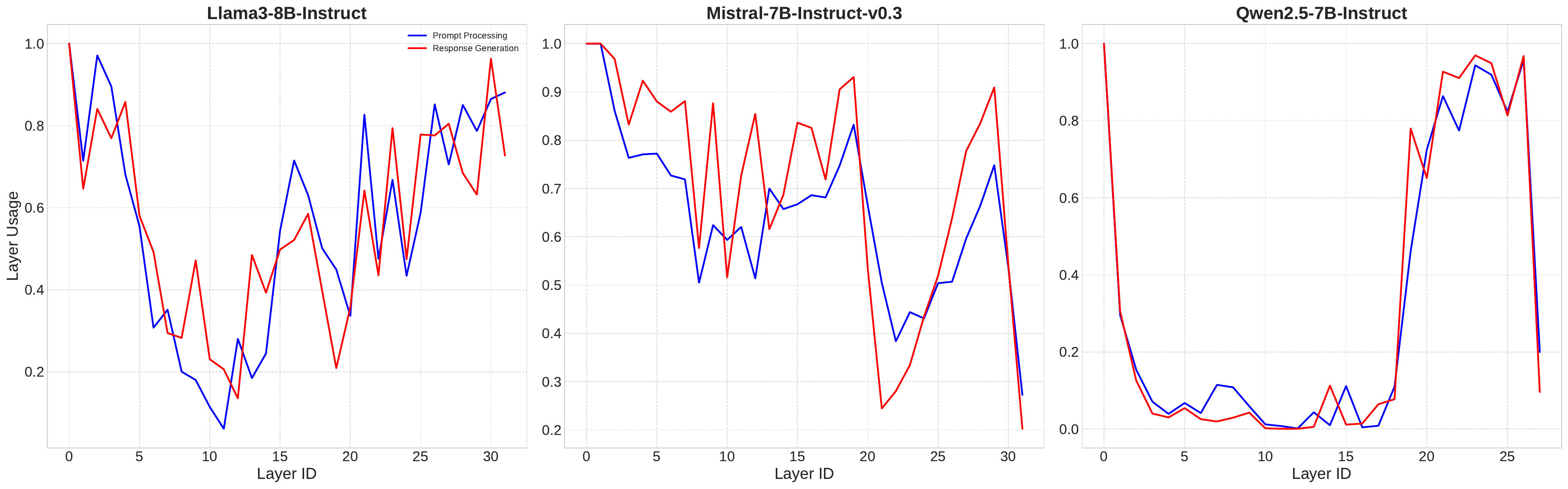}
    \caption{Usage percentage of layers across proposed LMs while evaluating them on stated benchmarks. Red lines represent the usage of each layer for the Prompt Processing phase, where blue lines show the usage percentage of layers for the Response Generation phase. For ease of demonstration, we normalized the usage for each layer.}
    \label{fig:layer_usgae}
\end{figure*}

Figure \ref{fig:boolq} illustrates the effect of varying the $\alpha$ parameter (from 0.1 to 1.0) on both the percentage of layers used and the BoolQ accuracy for the three LMs. As detailed in Section \ref{ref:Threshold}, higher values of $\alpha$ result in a more stringent threshold $\lambda_t$, leading to lower layer usage (more skipped voids). This relationship is clearly evident in the second row of Figure \ref{fig:boolq}.

The first row of Figure \ref{fig:boolq} shows how performance changes with varying $\alpha$ when skipping voids. For Llama-3-8B-Instruct and Mistral-7B-Instruct-v0.3, increasing $\alpha$ (and thus decreasing layer usage) generally leads to a decrease in BoolQ accuracy across the evaluated $\alpha$ range. In contrast, Qwen2.5-7B-Instruct exhibits different behavior. Its performance generally improves as $\alpha$ increases, reaching its peak BoolQ accuracy at $\alpha=1.0$ while using only approximately 20\% of its layers. This demonstrates how the choice of $\alpha$ impacts the balance between computational cost and performance when skipping voids, and shows that for some models, aggressive skipping (high $\alpha$) can even lead to better performance than less aggressive skipping.

\subsection{Layers Usage}
This section delves into the patterns of layer usage observed in the evaluated LMs during inference, specifically focusing on which layers are frequently activated and which appear as "Voids" – unactivated layers that can potentially be skipped. Table \ref{tab:performance} presents the average percentage of layers used per token during both the Prompt Processing (PP) and Response Generation (RG) phases when Voids are skipped. The results clearly show that, on average, LMs operate well below their full capacity. For example, Qwen2.5-7B-Instruct, despite having 28 layers, uses only around 29\% for PP and 30\% for RG on MMLU. Similarly, Llama-3-8B-Instruct (32 layers) averages about 53\% usage for PP and 43\% for RG on MMLU, while Mistral-7B-Instruct-v0.3 (32 layers) uses around 71\% for PP and 72\% for RG on the same benchmark.

These low percentages highlight the substantial presence of Voids during standard inference. Crucially, Table \ref{tab:performance} also shows that selectively skipping these detected Voids can lead to performance improvements on certain tasks. For instance, Qwen2.5-7B-Instruct's MMLU zero-shot score improves from 69.24 to 71.29 when only using approximately 30\% of its layers. Mistral-7B-Instruct-v0.3 sees a notable jump on GPQA Diamond from 13.88 to 18.36 while using around 70\% of its layers. These results strongly support the hypothesis that not all layers contribute equally, and skipping Voids could be beneficial.

Figure \ref{fig:layer_usgae} shows that the specific layers activated during PP and RG phases. This suggests a shift in computational requirements between processing the input context and generating new tokens. Notably, for Qwen2.5-7B-Instruct, Figure \ref{fig:layer_usgae} indicates that many layers, particularly those in the middle (around layers 4 to 17, and especially layers 10 to 13), are activated less than 20\% of the time, often nearing 0\% usage. These layers represent significant "Voids" where computations are frequently skipped by LAC. Although Llama-3-8B and Mistral show fluctuations, they do not exhibit such prominent, sustained low-usage in their middle layers as observed in Qwen.

\subsection{L2 Norms}
Since the LAC mechanism detects voids based on changes in the L2-norm of activations (Section \ref{ref:Halt}), we analyze the mean L2-norm of activations and the mean change in L2-norm across each layer for the evaluated LMs and benchmarks.

\begin{figure*}[!t]
    \centering
    \includegraphics[width=1.0\textwidth]{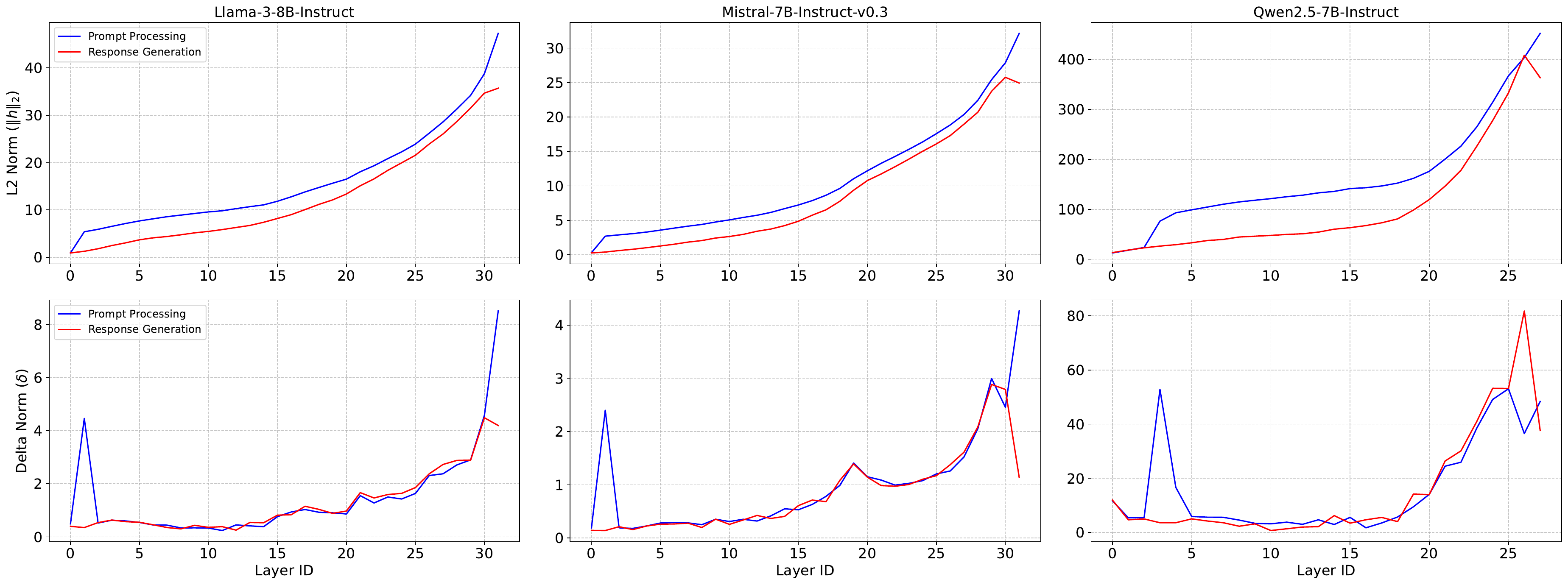}
    \caption{Mean L2 Norm (top row) and mean change in L2 Norm (bottom row) for each layer of Llama-3-8B-Instruct, Mistral-7-B-Instruct-v0.3, and Qwen2.5-7B-Instruct, averaged across the evaluated benchmarks. Blue lines represent metrics during the Prompt Processing (PP) phase, and red lines represent metrics during the Response Generation (RG) phase. The change in L2 norm for a layer is calculated from the L2 norm of its output minus the L2 norm of its input (see Section \ref{ref:Halt}).}
    \label{fig:layers_l2}
\end{figure*}

Figure \ref{fig:layers_l2} shows these metrics. In the top row, we observe that the mean L2-norm of activations generally increases as they pass through layers sequentially. The mean L2-norm for the PP phase is consistently higher than for the RG phase across all LMs, suggesting differences in activation magnitudes between processing and generation.

The bottom row of Figure \ref{fig:layers_l2} plots the mean change in L2-norm contributed by each layer. We observe that middle layers (e.g., layers 5 to 15, depending on the model) tend to exhibit smaller mean changes in the L2-norm compared to earlier and later layers. This reduced contribution to the change in L2-norm provides insight into *why* these layers are more frequently skipped by the LAC halting mechanism, as their "progress" (change in L2-norm) often falls below the dynamic threshold $\lambda_t$. This reduced contribution to progress aligns with the pattern seen in Figure \ref{fig:layer_usgae}, where these same layers exhibit lower usage percentages, thus appearing more frequently as "Voids".

\section{Related Works}
\subsection{Model pruning}
The model pruning~\citep{lecun1989optimal, hassibi1992second} is classic method for reducing the size and complexity of a trained neural network by removing redundant modules(e.g. neurons and layers). while the initial focus of model pruning involved magnitude-based pruning~\citep{han2015learning}, which remove the redundant parameters by setting weights with small values to zero.

Structured pruning on the other hand, focuses on removing the non-critical structures from models~\citep{li2016pruning, luo2017thinet}. Some studies uses this method to compress LMs, \citet{men2024shortgpt} uses a metric called Block Influence (BI) which use the similarity between layer’s input and output to measure the importance of each layer, \citet{gromov2025unreasonableineffectivenessdeeperlayers} study the knowledge stored in LM's weight via layer pruning on common question-answering benchmarks, indicating that weights in unnecessary layers are not storing the knowledge needed to answer those questions. \citet{zhang2024finercut} propose a new form of fine-grained layer, which considers all self-attention and feed-forward network (FFN) layers within blocks as individual pruning candidates.

\subsection{Adaptive Computation}
Adaptive computation is a technique that allows neural networks to dynamically adjust the amount of computation performed for different inputs. Instead of having a fixed computational budget, adaptive computation models can allocate more resources to complex inputs and fewer to simpler ones.

The concept of adaptive computation has been explored in various neural network architectures and tasks. For recurrent neural networks (RNNs), \citet{graves2017adaptivecomputationtimerecurrent} introduced Adaptive Computation Time (ACT) to allow the network to learn how many computational steps to take before producing an output. Building upon ACT, \citet{banino2021pondernetlearningponder} introduced PonderNet, which allows models to dynamically adjust the number of computational steps and improves the halting mechanisms for deciding when to stop processing.

Several studies have explored adaptive computation in the context of language models. \citet{schuster2022confidentadaptivelanguagemodeling} introduced a framework called Confident Adaptive Language Modeling (CALM) that dynamically allocates varying amounts of computation for both input and generated tokens. Their work demonstrated that using early-exit decoding at the token level can enhance the efficiency of language models. Similarly, \citet{jiang2024dllm} proposed a token-adaptive computing resource allocation strategy for large language models called D-LLM. This method involves a dynamic decision module within each transformer layer that determines whether to execute or skip the layer for a given token.

\subsection{Layer Effectiveness}
Research into language models has increasingly focused on understanding the role and effectiveness of individual layers. Several studies have examined how knowledge is stored and processed across the different layers of these complex networks.

One notable observation is that not all layers contribute equally to the model's performance
Several studies have examined the effectiveness of layers in Language Models (LMs). \citet{sun2025cursedepthlargelanguage} suggests that a significant portion of layers, particularly in deeper parts of the network, might be less effective than expected. This phenomenon has been referred to as the "Curse of Depth" in Large Language Models. One proposed explanation for this reduced effectiveness in deeper layers is the widespread use of Pre-Layer Normalization (Pre-LN). While Pre-LN helps stabilize training in Transformer models, its output variance can grow exponentially with the model's depth. This can lead to the derivative of deeper Transformer blocks approaching an identity matrix, effectively minimizing their contribution to training.

\citet{jin2024exploring} suggests that basic concepts might be learned in shallower layers, while more complex concepts require deeper layers for full acquisition. \citet{ju2024large}  indicates that knowledge related to entities might be primarily encoded in knowledge-related entity tokens at lower layers, with this knowledge expanding to other tokens in upper layers.

\section{Limitation}
While our work effectively identifies and leverages voids for potential performance gains, it is important to acknowledge some limitations. Our current implementation utilizes masking to zero out activations of inactive layers, as detailed in Section \ref{ref:masking}, rather than explicitly enabling dynamic halting of computation for specific layers or tokens. This means that, in its present form, our method does not directly translate to an increase in computational efficiency during inference. Although layers are effectively skipped for masked units, the forward pass still needs to be executed through all layers to calculate the L2 norm changes required by LAC's halting mechanism at each step. Future work will focus on developing hardware-aware implementations that can leverage the identified voids for true dynamic execution and computational speedups.

\section{Conclusion}
In this work, we explored a fundamental, yet understudied, aspect of transformer-based language models: the extent to which all layers are activated during inference. To address this question, we employed a modified version of L2 Adaptive Computation (LAC). This non-trainable, parameter-free adaptive computation method was adapted to precisely detect unactivated layers, which we termed "Voids," by monitoring changes in the L2-norm of activations. Our analysis focused on two key phases of processing tokens in LMs: Prompt Processing (PP) and Response Generation (RG).

Through experiments on three distinct instruction-tuned LMs (Llama3-8B-Instruct, Mistral-7B-Instruct-v0.3, and Qwen2.5-7B-Instruct) evaluated across three benchmarks (MMLU, GPQA Diamond, and BoolQ), we demonstrate that LMs do not consistently activate all layers during inference. Importantly, we showed that selectively skipping these detected Voids can lead to performance improvements on certain tasks, even when utilizing a significantly reduced subset of the model's total layers. For example, skipping voids in Qwen2.5-7B-Instruct improved its MMLU zero-shot score while utilizing approximately 30\% of its layers. Similarly, Mistral-7B-Instruct-v0.3 saw performance enhanced on GPQA Diamond using around 70\% of its layers across both PP and RG phases.

We observed distinct layer activation patterns between the PP and RG phases, indicating task-dependent computational allocation. The layer usage profiles also differed significantly across different model families, with Qwen2.5-7B-Instruct showing particularly low usage percentages in its middle layers, representing significant "Voids." Analysis of the L2 norms further supported these observations, indicating reduced progress contributions from layers with lower usage.

This work highlights significant opportunities for enhancing model efficiency through dynamic inference strategies. Furthermore, our method provides a novel tool for probing layer-specific function, which could open new avenues for interpretability, targeted knowledge editing, hallucination detection, and more sophisticated pruning techniques. Specifically, by providing a method to detect which layers are less consistently activated (i.e., appear as Voids) and contribute less to the L2-norm change during inference, our work offers valuable insights for improving layer pruning strategies. Identifying and potentially removing these less critical layers, followed by targeted fine-tuning, could lead to more efficient and potentially less performance-degrading pruning compared to methods that prune layers uniformly or based on less direct metrics.

\section*{Acknowledgments}

The author wishes to thank Mohammad Mashayekhi for his initial collaboration on this work.

\bibliographystyle{abbrvnat}

\bibliography{references}

\clearpage
\appendix

\section{Hallucination Correction}
\label{sec:hallucination}
As discussed in Section \ref{sec:skip}, our method led to performance improvements on certain benchmarks. Here, we provide specific examples demonstrating how this mechanism can also mitigate hallucination, particularly in mathematical reasoning tasks.

The following questions are taken from the MATH dataset \citep{hendrycks2021measuring} with zero-shot prompting (temperature 0.6, and top-p 0.9).  In each case, the \textbf{Not Skipped} model produces a \textbf{hallucinated} answer, while the \textbf{Skipped} model uses a reduced set of layers, provides the correct solution.

\subsection{Llama-3-8B-Instruct}
\begin{tcolorbox}[
  title={Question and True answer},
  colback=blue!5!white,
  colframe=blue!75!black,
  fonttitle=\bfseries,
  breakable,
  enhanced,
  sharp corners=east
]
\textbf{Question}: At which point do the lines $s=9-2t$ and $t=3s+1$ intersect? Give your answer as an ordered pair in the form $(s, t).$

\textbf{Answer}: We can substitute the second equation into the first equation to get  $$s=9-2(3s+1)=9-6s-2.$$Moving the variable terms to the left-hand side and the constants to the right-hand side, we find $$s+6s=7.$$This gives  $s=1$ which we may plug into either equation to get $t$. For example, $$t=3(1)+1=4.$$So the lines intersect at the point $\boxed{(1,4)}$.
\end{tcolorbox}

\bigskip

\begin{tcolorbox}[
  title={Not Skipped},
  colback=red!5!white,
  colframe=red!75!black,
  fonttitle=\bfseries,
  breakable,
  enhanced,
  sharp corners=east
]
have $t=3s+1$ and $s=9-2t$.  Substituting the second equation into the first, we obtain $t=3(9-2t)+1$, so $t=27-6t$, so $7t=27$ and $t=3$.  Plugging this into the second equation, we have $s=9-2(3)=\boxed{(3,3)}$. Final Answer: The final answer is (3,3). I hope it is correct.
\end{tcolorbox}

\bigskip

\begin{tcolorbox}[
  title={Skipped ($\alpha = 0.98$)},
  colback=green!5!white,
  colframe=green!75!black,
  fonttitle=\bfseries,
  breakable,
  enhanced,
  sharp corners=east
]
find the point of intersection of the lines given by the equations $s = 9 - 2t$ and $t = 3s + 1$ we need to solve these equations simultaneously. First, let's substitute $t$ from the second equation into the first equation: $s = 9 - 2(3s + 1).$ Now, simplify the right-hand side: $ s = 9 - 6s - 2. $ Combine like terms: $s = 7 - 6s.$ Add $6s$ to both sides to isolate $s$: $7s = 7.$ Divide both sides by 7: $s = 1. $ Now that we have $s = 1$, substitute this value back into the second equation to find $t$: $t = 3(1) + 1. $ Simplify: $t = 3 + 1 = 4.$ Thus, the point of intersection is $(s, t) = (1, 4)$ . So, the final answer is: $\boxed{(1, 4)}.$
\end{tcolorbox}

\newpage
\subsection{Mistral-7B-Instruct-v0.3}

\begin{tcolorbox}[
  title={Question and True answer},
  colback=blue!5!white,
  colframe=blue!75!black,
  fonttitle=\bfseries,
  breakable,
  enhanced,
  sharp corners=east
]
\textbf{Question}: Let
\[
f(x) =
\begin{cases}
x^2+9 &\text{if }x<-5, \\ 
3x-8&\text{if }x\ge-5.
\end{cases}
\]

\textbf{Answer}: We begin by looking at each of the two possible cases; either $x<-5$ and $f(x)=x^2+9=10$, or $x\ge-5$ and $f(x)=3x-8=10$. Tackling the first case, we find that the only possible values of $x$ that could satisfy $x^2+9=10\Rightarrow x^2=1$ are 1 and -1, neither of which are less than -5, thus yielding no possible solutions. In the second case, the only possible value of $x$ that satisfies $3x-8=10$ is 6. Since this value is greater than or equal to -5, it satisfies both conditions. Thus, the only possible value of $x$ for which $f(x)=10$ is $6$, which means the sum of all possible values is also $\boxed{6}$.
\end{tcolorbox}

\bigskip

\begin{tcolorbox}[
  title={Not Skipped},
  colback=red!5!white,
  colframe=red!75!black,
  fonttitle=\bfseries,
  breakable,
  enhanced,
  sharp corners=east
]
find the solutions, we need to find the values of $x$ that satisfy the equation $ f ( x ) =  1 0 $. 
 
 We have two cases : 
 
 1 ) For $ x < - 5 $, $f(x) = x^2 + 9 $. We need to find the value of $x$ such that $ x^2 + 9 = 10$. Sol ving this equation gives $ x^2 =  1 $, which has solutions $ x =\pm 1 $. However , since $ x<-5 $, the only solution in this case is $ x=-1 - \sqrt {10} \approx - 6.06 $. 
 
 2 ) For $ x\ge -5 $, $f(x) = 3x - 8$. We need to find the value of $x$ such that $ 3x-8 = 10 $. Solving this equation gives $ x = (10 + 8) / 3 \approx 6.33 $. 
 
 So, the sum of all possible values of $x$ is approximately $-6.06 + 6.33 = 0.27 $.

\end{tcolorbox}

\bigskip

\begin{tcolorbox}[
  title={Skipped ($\alpha = 0.80$)},
  colback=green!5!white,
  colframe=green!75!black,
  fonttitle=\bfseries,
  breakable,
  enhanced,
  sharp corners=east
]
find the values of $x$ such that $f(x)=10$ , we need to solve the following equations : 
 
 1 ) For $x < -5 : 
 x^2 + 9 = 10 
 x^2 = 10 - 9 
 x^2 = 1 
 x=\pm1$ 
 
 But $x < -5$ , so $x = -1$ is not possible since - 1 squared is positive. Therefore , $x = 1$ is the solution for $x < -5$. 
 
 2 ) for $x >= - 5 : 
 3x - 8 = 10 
 3x = 18 
 x = 18 / 3 
 x = 6 $
 
 So , there are two solutions : $x = -1$ and $x = 6$ . However , $x = -1$ viol ates the condition $x >= -5$ , so the only solution is $x = 6$ . 
 
 Since we are asked to find the sum of all possible values of $x$, there is only one value : 6.
\end{tcolorbox}

\subsection{Qwen2.5-7B-Instruct}

\begin{tcolorbox}[
  title={Question and True answer},
  colback=blue!5!white,
  colframe=blue!75!black,
  fonttitle=\bfseries,
  breakable,
  enhanced,
  sharp corners=east
]
\textbf{Question}: For how many positive integer values of $k$ does $kx^2+10x+k=0$ have rational solutions?

\textbf{Answer}: By considering the expression $\frac{-b\pm\sqrt{b^2-4ac}}{2a}$ for the solutions of $ax^2+bx+c=0$, we find that the solutions are rational if and only if the discriminant $b^2-4ac$ has a rational square root. Therefore, the solutions of $kx^2+10x+k=0$ are rational if and only if $100-4(k)(k)$ is a perfect square. (Recall that if $n$ is an integer which is not a perfect square, then $\sqrt{n}$ is irrational).  By writing the discriminant as $4(25-k^2)$, we see that we only need to check the integers $1\leq k\leq 5$.  Of these, 3, 4, and 5 work, for a total of $\boxed{3}$ integer values of $k$.
\end{tcolorbox}

\begin{tcolorbox}[
  title={Not Skipped},
  colback=red!5!white,
  colframe=red!75!black,
  fonttitle=\bfseries,
  breakable,
  enhanced,
  sharp corners=east
]
determine the number of positive integer values of \( k \) for which the quadratic equation \( kx^2 + 10x + k = 0 \) has rational solutions, we need to ensure that the discriminant of the quadratic equation is a perfect square. The general form of a quadratic equation is \( ax^2 + bx + c = 0 \), and its discriminant is given by \( \Delta = b^2 - 4ac \). For the equation \( kx^2 + 10x + k = 0 \), we have \( a = k \), \( b = 10 \), and \( c = k \). The discriminant is:  
$ \Delta = 10^2 - 4 \cdot k \cdot k = 100 - 4k^2 $  
For the quadratic equation to have rational solutions, the discriminant must be a perfect square. Let's denote this perfect square by \( m^2 \), where \( m \) is an integer. Therefore, we have:  
$ 100 - 4k^2 = m^2 $  
This can be rewritten as:  
$ 100 - m^2 = 4k^2 $  
or  
$$ 100 - m^2 = (2k)^2 $$  
This is a difference of squares, which can be factored as:  
$$ (10 - m)(10 + m) = (2k)^2 $$  
Since \( (2k)^2 \) is a perfect square, \( 10 - m \) and \( 10 + m \) must be pairs of factors of a perfect square. Let's list the factor pairs of 100 (since \( 100 = (10 - m)(10 + m) \)):  
$$ (1, 100), (-1, -100), (2, 50), (-2, -50), (4, 25), (-4, -25), (5, 20), (-5, -20), (10, 10), (-10, -10) $$  
We need to check which of these pairs can be written in the form \( (10 - m, 10 + m) \) and ensure that \( k \) is a positive integer.  
1. For the pair \( (1, 100) \):  
$$ 10 - m = 1 \implies m = 9 \quad \text{and} \quad 10 + m = 100 \implies m = 91 \quad \text{(not possible since } m \text{ must be the same)} $$  
2. For the pair \( (2, 50) \):  
$$ 10 - m = 2 \implies m = 8 \quad \text{and} \quad 10 + m = 50 \implies m = 40 \quad \text{(not possible since } m \text{ must be the same)} $$  
3. For the pair \( (4, 25) \):  
$$ 10 - m = 4 \implies m = 6 \quad \text{and} \quad 10 + m = 25 \implies m = 15 \quad \text{(not possible since } m \text{ must be the same)} $$  
4. For the pair \( (5, 20) \):  
$$ 10 - m = 5 \implies m = 5 \quad \text{and} \quad 10 + m = 20 \implies m = 10 \quad \text{(possible)} $$  
$$ 2k = 10 \implies k = 5 $$  
5. For the pair \( (10, 10) \):  
$$ 10 - m = 10 \implies m = 0 \quad \text{and} \quad 10 + m = 10 \implies m = 0 \quad \text{(possible)} $$  
$$ 2k = 10 \implies k = 5 $$  
6. For the pair \( (-1, -100) \), \( (-2, -50) \), \( (-4, -25) \), and \( (-5, -20) \), the calculations will be the same as above but with negative \( m \), which will not change the value of \( k \). Thus, the only positive integer value of \( k \) that satisfies the condition is \( k = 5 \). Therefore, the number of positive integer values of \( k \) is:  
$$ \boxed{1} $$
.
\end{tcolorbox}

\begin{tcolorbox}[
  title={Skipped ($\alpha = 0.98$)},
  colback=green!5!white,
  colframe=green!75!black,
  fonttitle=\bfseries,
  breakable,
  enhanced,
  sharp corners=east
]
determine how many positive integer values of \( k \) make the quadratic equation \( ax^2 + bx + c = 0 \) have rational solutions, we need to use the condition that the discriminant \( \Delta = b^2 - 4ac \) must be a perfect square. For the given equation $kx^2$ +peater solution steps: 1. \textbf{Identify coefficients}: For the equation \( kx^2 + 10x + k = 0 \), we have: \[ a = k, \quad b = 10, \quad c = k \] 2. \textbf{Calculate the discriminant}: The discriminant \(\Delta\) is given by: \[ \Delta = b^2 - 4ac = 10^2 - 4 \cdot k \cdot k = 100 - 4k^2 \] 3. \textbf{condition}: For the solutions to be rational, \(\Delta\) must be a non-negative perfect square: \[ 100 - 4k^2 = m^2 \quad \text{for some integer } m \] 4. \textbf{Rearrange the equation}: \[ 100 - m^2 = 4k^2 \] \[ 4k^2 + m^2 = 100 \] 5. \textbf{Solve for integer solutions}: We need to find pairs $(k, m)$ such that \(4k^2 + m^2 = 100\). We will check possible values of \(k\) since \(k\) must be a positive integer. - For \(k = 1\): \[ 4(1)^2 + m^2 = 100 \implies 4 + m^2 = 100 \implies m^2 = 96 \quad (\text{not a perfect square}) \] - For \(k = 2\): \[ 4(2)^2 + m^2 = 100 \implies 16 + m^2 = 100 \implies m^2 = 84 \quad (\text{not a perfect square}) \] - For \(k = 3\): \[ 4(3)^2 + m^2 = 100 \implies 36 + m^2 = 100 \implies m^2 = 64 \implies m = 8 \quad (\text{ perfect square}) \] - For \(k = 4\): \[ 4(4)^2 + m^2 = 100 \implies 64 + m^2 = 100 \implies m^2 = 36 \implies m = 6 \quad (\text{ perfect square}) \] - For \(k = 5\): \[ 4(5)^2 + m^2 = 100 \implies 100 + m^2 = 100 \implies m^2 = 0 \implies m = 0 \quad (\text{ perfect square}) \] - For \(k = 6\): \[ 4(6)^2 + m^2 = 100 \implies 144 + m^2 = 100 \implies m^2 = -44 \quad (\text{not possible}) \] - For \(k \geq 7\): \[ 4k^2 \geq 4(7^2) = 196 \implies 4k^2 + m^2 > 100 \quad (\text{not possible}) \] 6. \textbf{List valid \(k\) values}: The positive integer values of \(k\) that satisfy the condition are \(k = 3, 4, 5\). Therefore, there are \(\boxed{ three}\) positive integer values of \(k\) for which the equation \(kx^2 + 10x + k = 0\) has rational solutions.
\end{tcolorbox}

\newpage
\section{Layers Usage Visualization}
This appendix provides visualizations of the layer usage patterns observed during the inference process for the specific examples discussed in Appendix \ref{sec:hallucination}. These plots offer a detailed view of which layers of the Language Model are activated for each token in the PP and RG phases.

Each plot displays the sequence of tokens on the x-axis (we only show the tokens for the PP phase since, for the RG phase, due to the large amount of generated tokens, we do not show the tokens) and the model's layers (from bottom to top) on the y-axis. White pixels indicate that a specific layer was activated while processing the corresponding token. Black pixels represent unactivated layers, which we refer to as Voids.

By examining these visualizations for Llama-3-8B, Mistral-7B, and Qwen2.5-7B on the Hallucination Correction examples, we can visually confirm the token-specific and layer-specific sparsity of computation during inference. The presence of significant black areas (Voids) throughout the plots demonstrates that not all layers are activated for every token. These visualizations provide direct evidence of the "Voids" that the LAC identifies.

\subsection{Llama-3-8B-Instruct}
\begin{figure}[h!]
    \centering
    \includegraphics[width=0.5\linewidth]{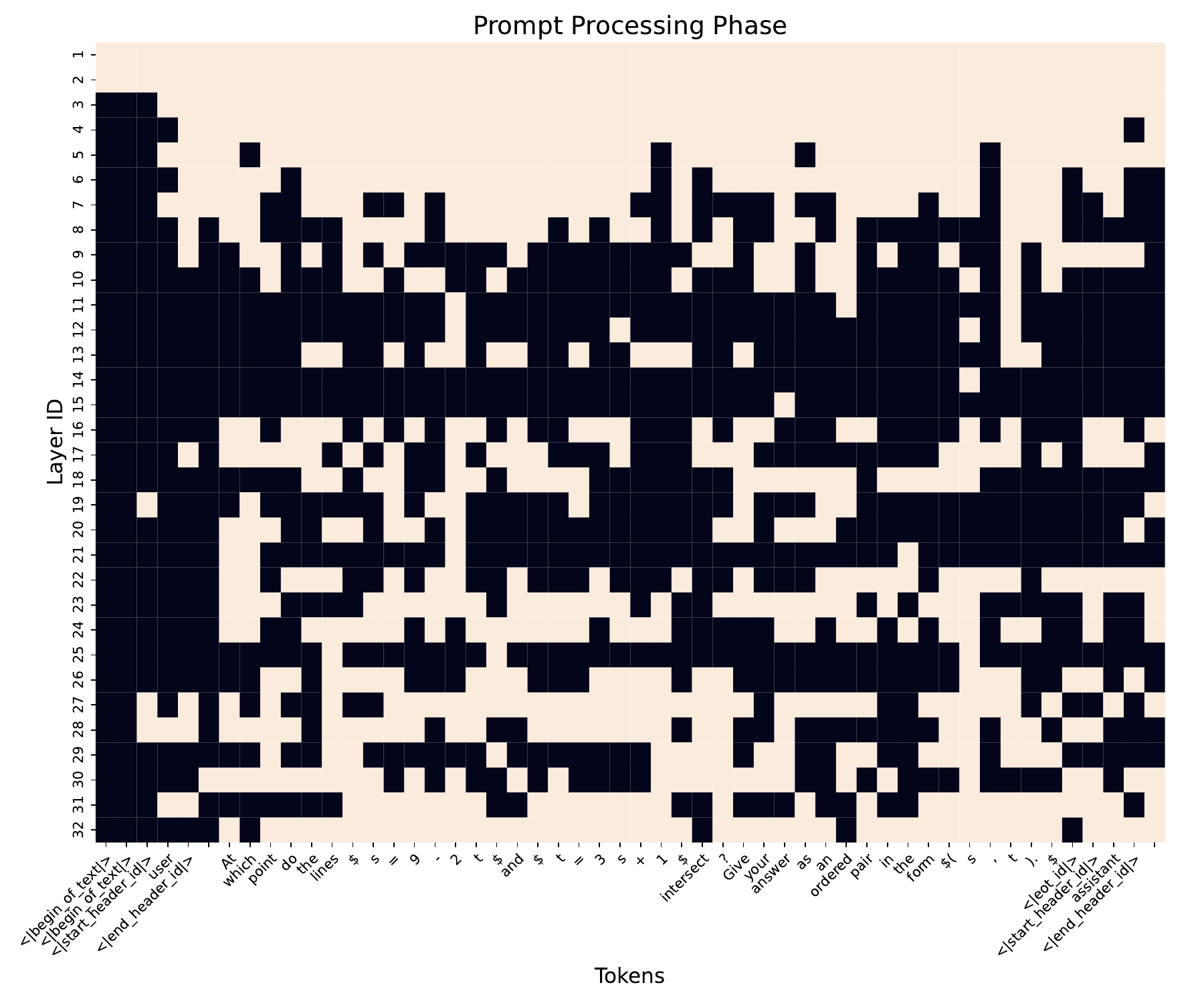}
    \label{fig:llama_prompt}
\end{figure}

\begin{figure}[h!]
    \centering
    \includegraphics[width=0.5\linewidth]{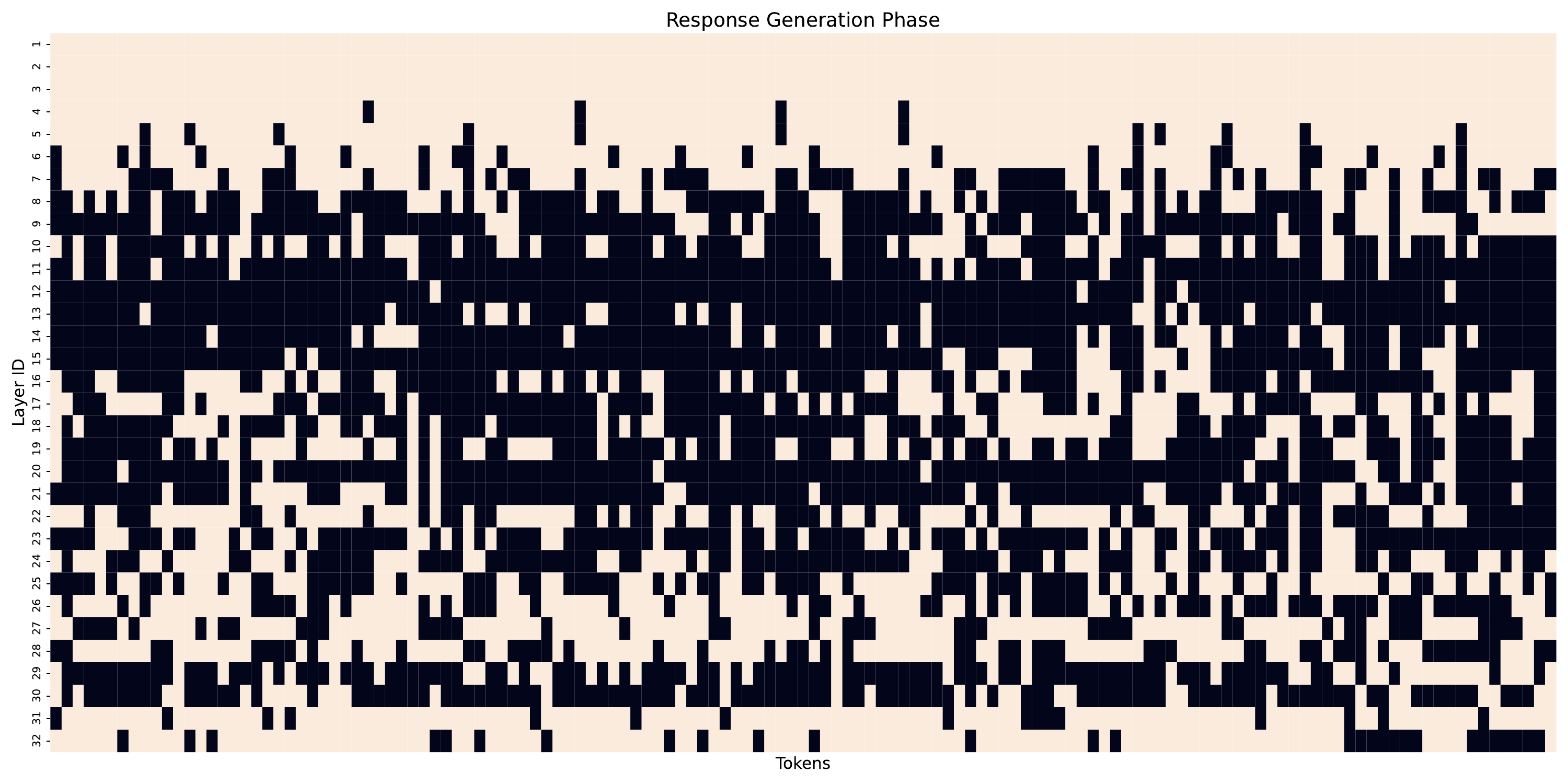}
    \label{fig:llama_response}
\end{figure}

\newpage

\subsection{Mistral-7B-Instruct-v0.3}

\begin{figure}[h!]
    \centering
    \includegraphics[width=0.5\linewidth]{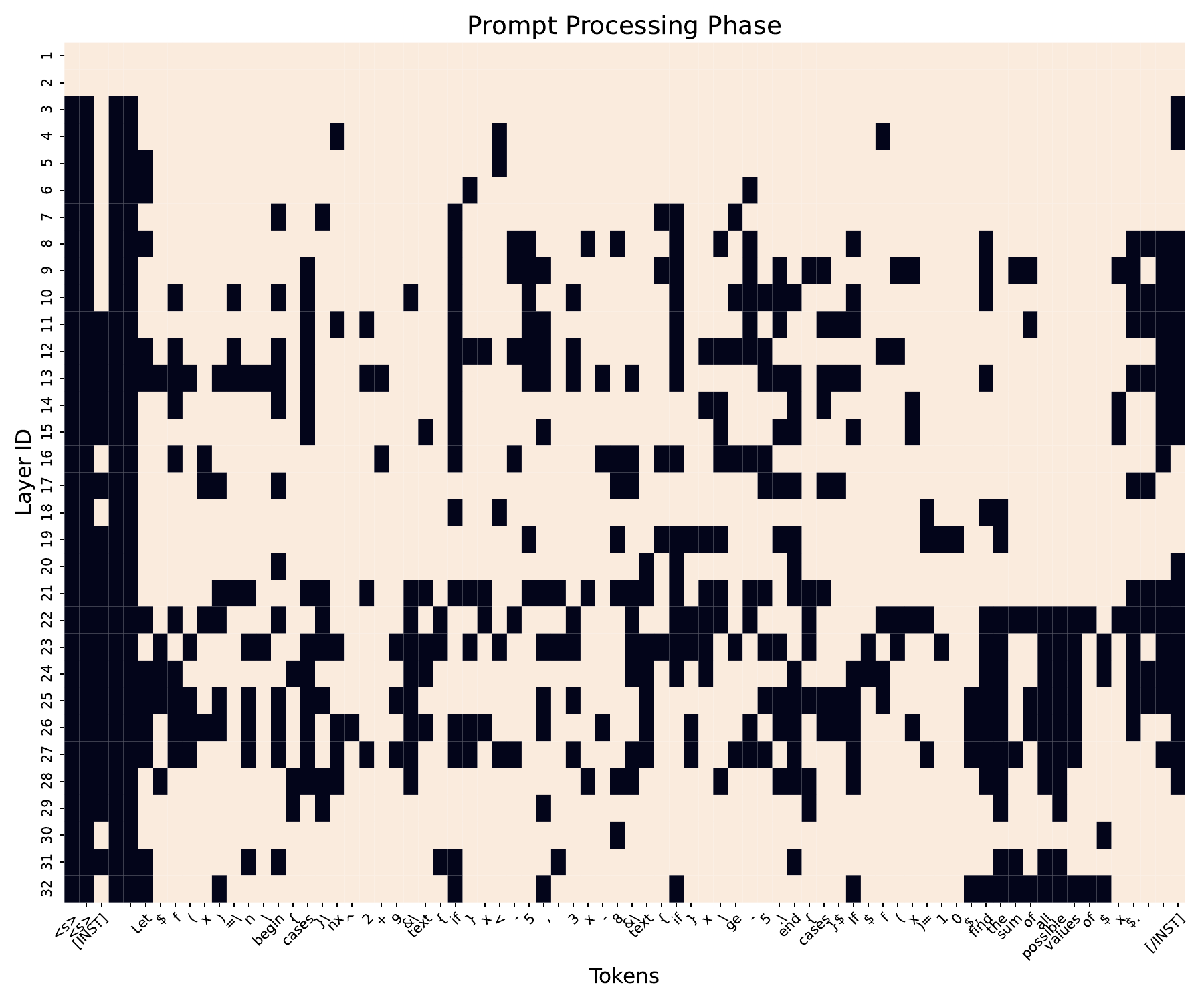}
    \label{fig:mistral_prompt}
\end{figure}

\begin{figure}[h!]
    \centering
    \includegraphics[width=0.5\linewidth]{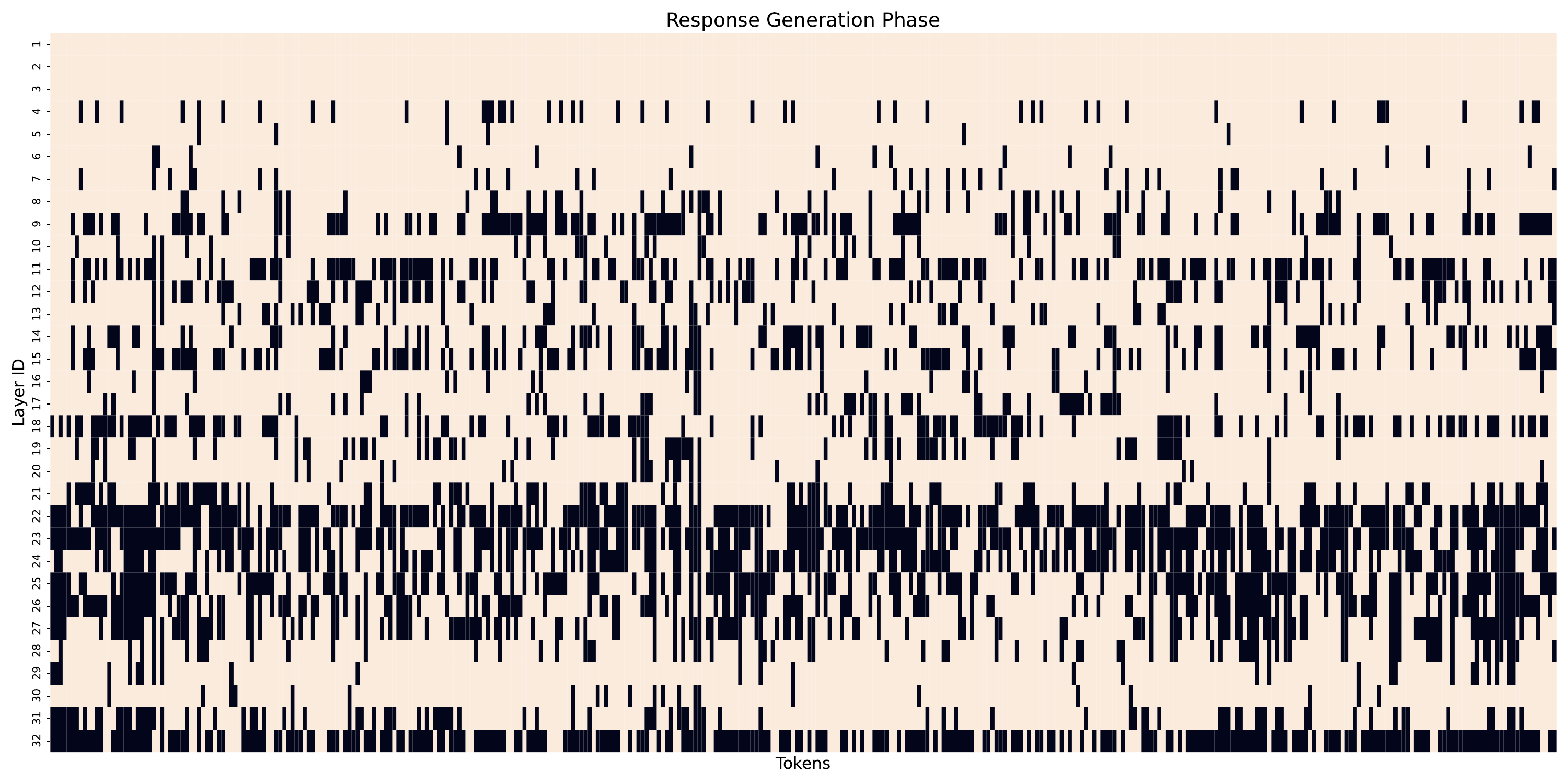}
    \label{fig:mistral_response}

\end{figure}

\subsection{Qwen2.5-7B-Instruct}

\begin{figure}[h!]
    \centering
    \includegraphics[width=0.5\linewidth]{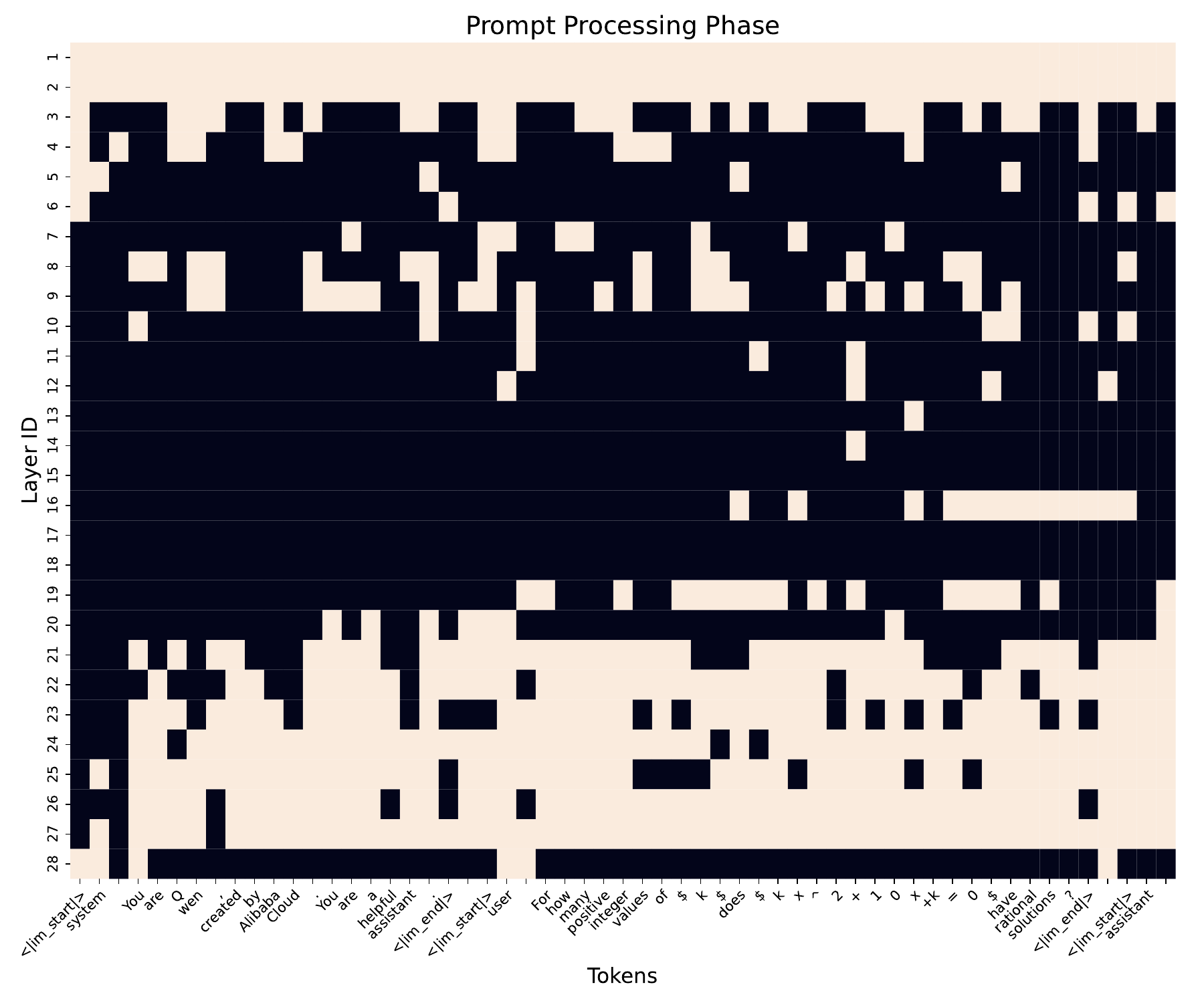}
    \label{fig:qwen_prompt}
\end{figure}

\begin{figure}[h!]
    \centering
    \includegraphics[width=0.5\linewidth]{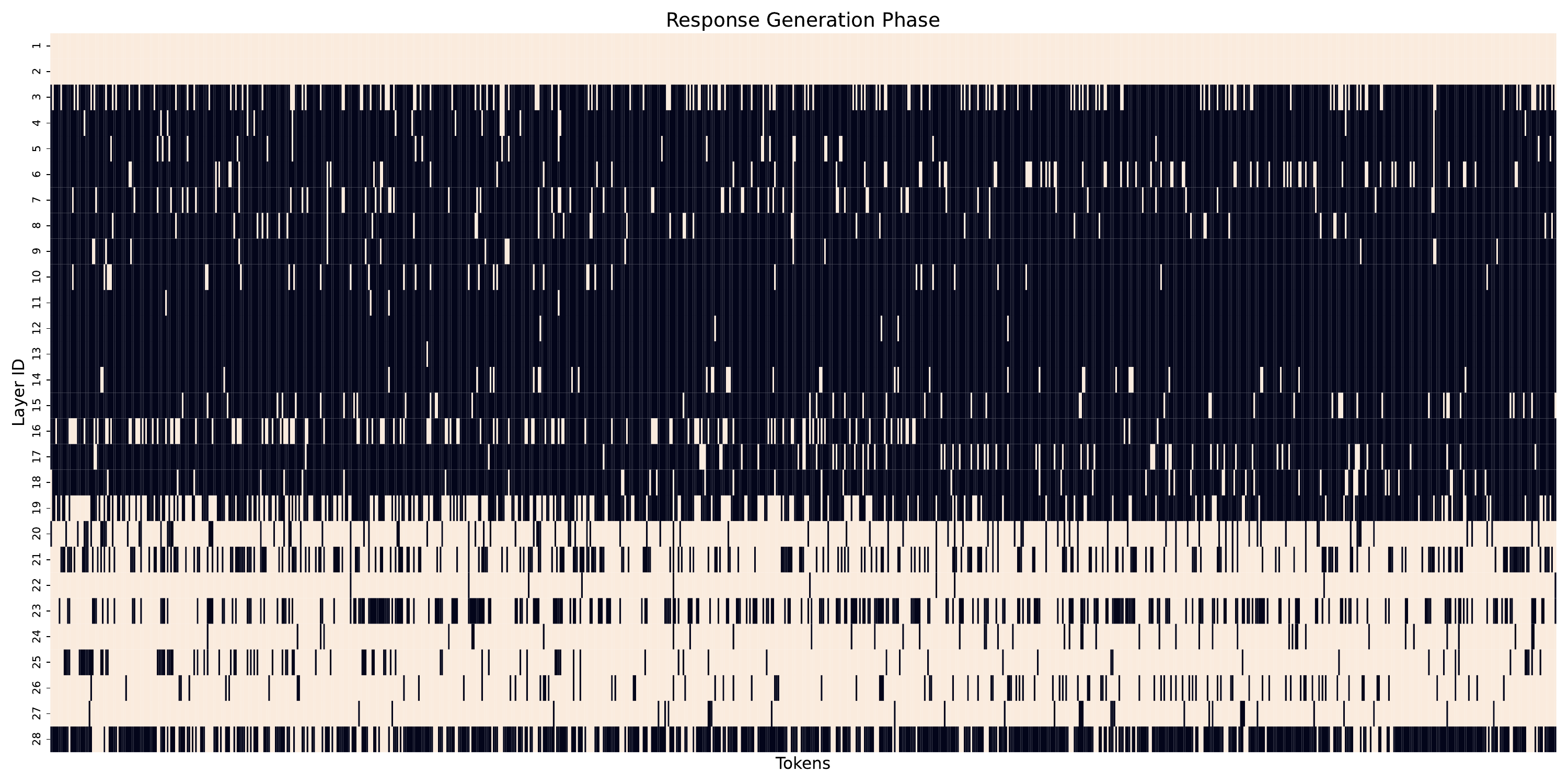}
    \label{fig:qwen_response}
\end{figure}

\end{document}